\PassOptionsToPackage{numbers, sort&compress}{natbib}
\documentclass{article}
\usepackage[final]{neurips_2024}

\usepackage[utf8]{inputenc} 
\usepackage[T1]{fontenc}    
\usepackage{hyperref}
\hypersetup{
  colorlinks=true,
  linkcolor=blue,
  citecolor=blue,
  urlcolor=black
}

\usepackage{url}                 
\usepackage{amsmath}
\usepackage{amsfonts}
\usepackage{amssymb}
\usepackage{nicefrac}
\usepackage{microtype}
\usepackage{xcolor}
\usepackage{graphicx}
\usepackage{multirow}
\usepackage{booktabs}
\usepackage{tabularx}
\usepackage{subcaption}
\usepackage{makecell}
\usepackage{enumitem}
\usepackage{adjustbox}
\usepackage{tcolorbox}
\usepackage{dialogue}
\usepackage{longtable}
\usepackage{colortbl}
\usepackage{float}

\usepackage[normalem]{ulem}
\useunder{\uline}{\ul}{}

\renewcommand{\thefootnote}{\fnsymbol{footnote}}

\title{SocioVerse: A World Model for Social Simulation Powered by LLM Agents and A Pool of 10 Million Real-World Users}

\author{
    Xinnong Zhang\textsuperscript{\rm 1,2}\footnotemark[2], 
    Jiayu Lin\textsuperscript{\rm 1,2}\footnotemark[2], 
    Xinyi Mou\textsuperscript{\rm 2}\footnotemark[2], 
    Shiyue Yang\textsuperscript{\rm 2}, 
    Xiawei Liu\textsuperscript{\rm 2}, 
    \\
    \textbf{
    Libo Sun\textsuperscript{\rm 2}, 
    Hanjia Lyu\textsuperscript{\rm 3}, 
    Yihang Yang\textsuperscript{\rm 2}, 
    Weihong Qi\textsuperscript{\rm 4}, 
    Yue Chen\textsuperscript{\rm 2},
    }\\
    \textbf{
    Guanying Li\textsuperscript{\rm 2},
    Ling Yan\textsuperscript{\rm 5},
    Yao Hu\textsuperscript{\rm 5},
    Siming Chen\textsuperscript{\rm 2},
    Yu Wang\textsuperscript{\rm 2},
    }\\
    \textbf{
    Xuanjing Huang\textsuperscript{\rm 2}, 
    Jiebo Luo\textsuperscript{\rm 3}, 
    Shiping Tang\textsuperscript{\rm 2}, 
    Libo Wu\textsuperscript{\rm 1,2}, 
    Baohua Zhou\textsuperscript{\rm 2}, 
    Zhongyu Wei\textsuperscript{\rm 1,2}
    }\\
    \normalsize\textsuperscript{\rm 1}{Shanghai Innovation Insititute}, 
    \normalsize\textsuperscript{\rm 2}{Fudan University},\\
    \normalsize\textsuperscript{\rm 3}{University of Rochester}, 
    \normalsize\textsuperscript{\rm 4}{Indiana University},
    \normalsize\textsuperscript{\rm 5}{Xiaohongshu Inc.}\\
    \normalsize\texttt{zywei@fudan.edu.cn}\\
    \hyperref[https://github.com/FudanDISC/SocioVerse]{\textbf{\textit{SocioVerse}}: \texttt{\textcolor{blue}{https://github.com/FudanDISC/SocioVerse}}}
    }

\begin{document}
\maketitle

\vspace{-1.2cm}
\begin{figure}[h!]
    \setlength{\abovecaptionskip}{-0.4cm}
    \centering
    \includegraphics[width=\columnwidth]{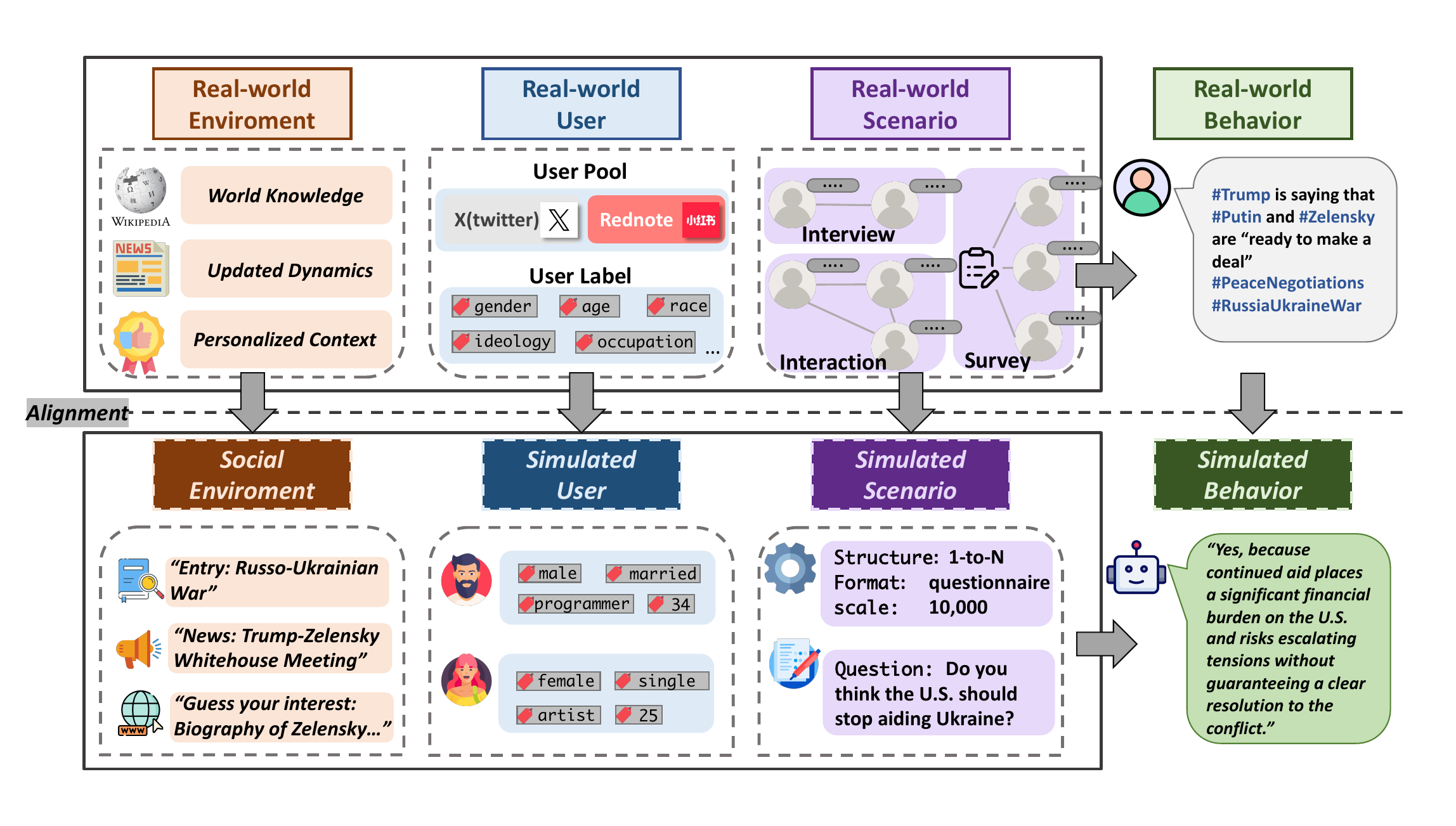}
    \caption{An illustration of the \textbf{\textit{SocioVerse}} in the case of Ukraine issue. The alignment challenges are well handled regarding environment, user, scenario, and behavior.}
    \label{fig:head_pic}
\end{figure}

\footnotetext[2]{These authors contribute equally to this work.}

\begin{abstract}

Social simulation is transforming traditional social science research by modeling human behavior through interactions between virtual individuals and their environments. With recent advances in large language models (LLMs), this approach has shown growing potential in capturing individual differences and predicting group behaviors. However, existing methods face alignment challenges related to the environment, target users, interaction mechanisms, and behavioral patterns. To this end, we introduce \textit{\textbf{SocioVerse}}, an LLM-agent-driven world model for social simulation. Our framework features four powerful alignment components and a user pool of 10 million real individuals. To validate its effectiveness, we conducted large-scale simulation experiments across three distinct domains: politics, news, and economics. Results demonstrate that SocioVerse can reflect large-scale population dynamics while ensuring diversity, credibility, and representativeness through standardized procedures and minimal manual adjustments.
\end{abstract}

\renewcommand{\thefootnote}{\arabic{footnote}}

\section{Introduction}\label{sec: intro}



The study of human behavior aims to understand how individuals and groups act in various social contexts and serves as a cornerstone of social science research. Traditionally, this has been accomplished using methods such as surveys, interviews, and observations~\cite{public-diplomacy, cologna2025trust, peisakhin2024hosts}. However, these approaches often encounter challenges, including high costs, limited sample sizes, and ethical concerns. As a result, researchers have resorted to alternative methods for studying human behavior.

Social simulation has emerged as an effective method for addressing this issue, where researchers use agents to model human behavior, observe their reactions, and translate these findings into insights about human behavior~\cite{schelling1971dynamic, smith2007agent}. By assigning behavioral rules to autonomous agents, researchers can explore how micro-level decisions lead to emergent macro-level patterns through the agent-based models~\cite{connolly1979micromotives, jackson2017agent}. This approach enables capturing specific groups' preferences on particular topics and forecasting potential social dynamics. Furthermore, recent advancements in large language models (LLMs) have significantly enhanced agents' reasoning and decision-making capabilities, enabling them to operate and interact within increasingly realistic and complex environments~\cite{argyle2023out, mou2024unveiling, mou2024unifying}.


Recent studies have explored social simulation across various levels and scenarios, from mimicking well-known individuals and mirroring specific situations to modeling large-scale social dynamics~\cite{mou2024individual,shao2023character,liu2024ai,bao2024piors,mou2024agentsensebenchmarkingsocialintelligence,yang2024oasis}. However, they share a common challenge: \textbf{alignment} between the simulated environment and the real world, which manifests across multiple dimensions and raises several key questions that remain to be addressed, as shown in Figure~\ref{fig:head_pic}.

\textbf{Q1. How to align the simulated environment with the real world?} \\
In the real world, new events occur every day and new content is continuously generated. The behavior of real users is rooted in these ever-evolving social contexts and policy agendas. However, the static knowledge of LLMs prevents them from aligning with the dynamic nature of the real-world social environment~\cite{gao2024largeq1, anthis2025llmq1}. There is a gap between the simulated context and the real world, which results in discrepancies between the simulation process and outcomes compared to those in reality. Therefore, it is necessary to establish an update mechanism to keep the simulated environment synchronized with the real world.



\textbf{Q2. How to align simulated agents with target users precisely?}\\
The composition of users in the real world is both complex and diverse, making it impractical to enumerate all users in every scenario. Therefore, it is essential to identify target users whose distribution aligns with that of the users in the corresponding scenario, thereby accurately reflecting the real-world composition and relationships~\cite{giorgi2022correcting, ribeiro2018media}. Based on this, precise target user simulation also requires providing agents with a detailed and comprehensive description of the corresponding users, often involving the integration of high-fidelity demographic, contextual, and behavioral data.




\textbf{Q3. How to align the interaction mechanism with the real world among different scenarios?}\\
The diversity of social interactions presents challenges in social simulation design, requiring deliberate choices regarding the number of individuals, social structures, interaction patterns, and message dissemination mechanisms, to align with the real world. This often results in independently constructed task-specific simulation pipelines performing repetitive work, which reduces their generalizability and scalability~\cite{lee2023can, xiao2023simulating}. Therefore, there is a need for unified simulation frameworks based on systematic categorization to standardize simulation components and facilitate extensibility across different social scenarios.



\textbf{Q4. How to align the behavioral pattern with the real-world groups?}\\
When the environment perceived by agents, the user composition, and the interaction mechanisms are aligned with the real world, agents are expected to exhibit responses consistent with those of the corresponding real users. However, current LLMs exhibit inherent bias and limitations in such reasoning, failing to infer different types of user behaviors~\cite{gao2023s3, yang2024oasis}. Therefore, it is necessary to systematically collect behavior-driving factors across different user characteristics and adopt appropriate modeling approaches to effectively capture diverse behavior patterns.

\begin{figure}
    \centering
    \includegraphics[width=\linewidth]{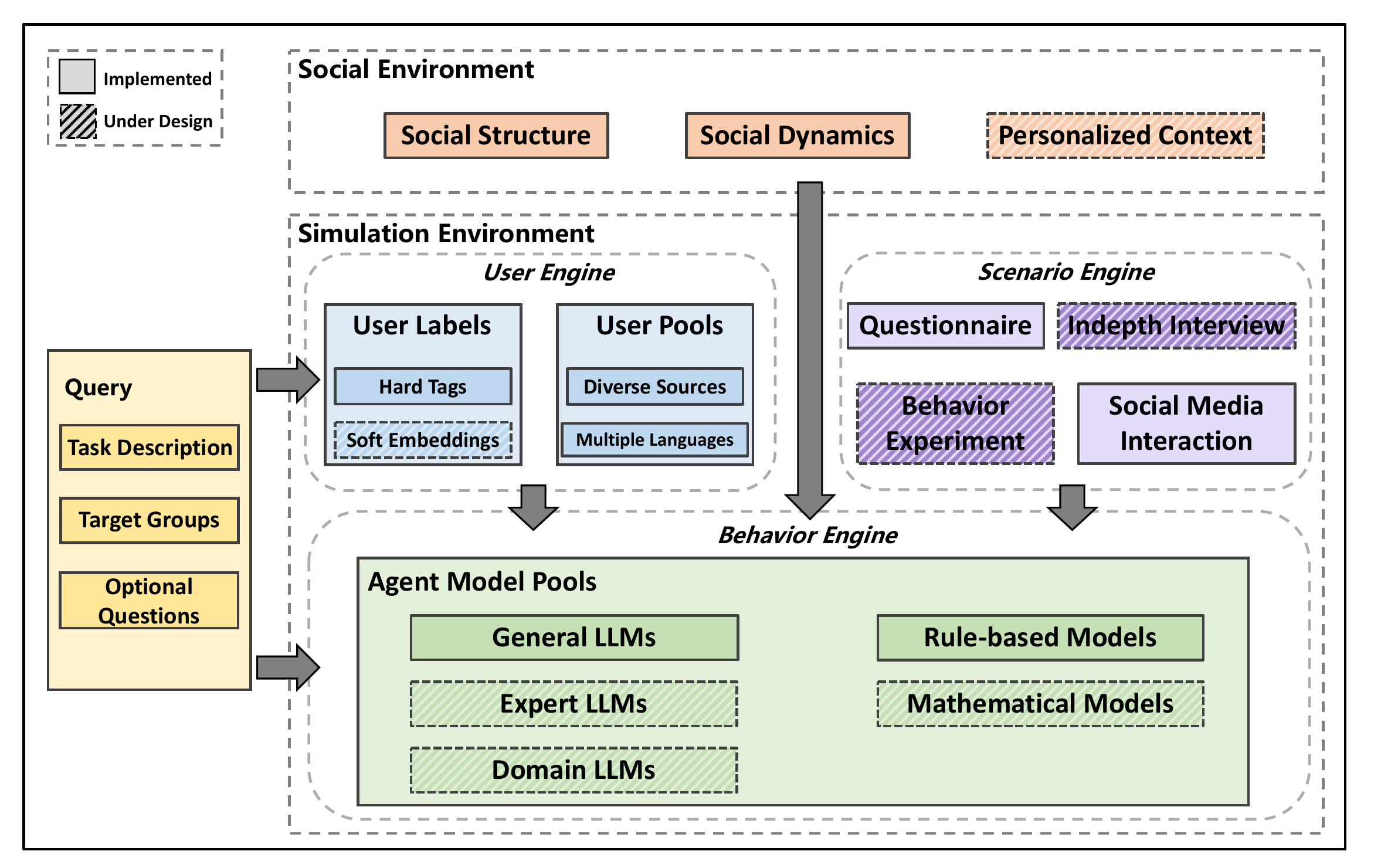}
    \caption{An illustration of \textit{\textbf{SocioVerse}} framework invovling 4 powerful parts. The social environment provides an updated context for the simulation. During the simulation, the behavior engine takes the simulation setting, user profiles, and social information from the scenario engine, user engine, and social environment, respectively, and generates the results according to the query.}
    \label{fig: framework}
\end{figure}

In this paper, we propose \textbf{\textit{SocioVerse}}, a world model for social simulation driven by LLM-based agents based on a large-scale real-world user pool. As shown in Figure~\ref{fig: framework}, we design modular components to address the above questions. The \textbf{Social Environment} injects up-to-date and external real-world information into the simulation. The \textbf{User Engine} and \textbf{Scenario Engine} respectively reconstruct realistic user context and orchestrate the simulation process to align the simulation with the real world. Given this rich contextual setup, the \textbf{Behavior Engine} then drives agents to reproduce human behaviors accordingly.

To support the framework, we construct a user pool of 10 million individuals by collecting real-world social media data to power the user engine. Comparable in scale to the entire populations of Hungary or Greece, this extensive pool enables diverse and large-scale social simulations. For any customized simulation task, various sampling strategies can be applied to extract target user groups from the pool to support the simulation process.


We conduct three simulations using the SocioVerse framework, each differing in research domain, user composition, and social environment:  (a) presidential election prediction, (b) breaking news feedback, and (c) national economic survey. For each task, we compare the simulation results with real-world situations. Extensive and comprehensive experiments demonstrate that our framework serves as a robust foundation for building standardized and accurate large-scale social simulations. In summary, our key contributions are as follows:

\begin{itemize}
    \item \textbf{SocioVerse}: We propose a world model for social simulation comprising four powerful alignment modules, enabling diverse and trustworthy social simulations (as illustrated in Figure~\ref{fig: framework}).
    \item \textbf{10M User Pool}: A user pool of 10 million individuals, constructed from real-world behavioral data, enables large-scale and diverse social simulations, ranging from small interest groups to large citizen communities.
    \item \textbf{Three Illustrative Simulations}: We demonstrate the framework's capabilities through three distinct scenarios: presidential election prediction, breaking news feedback, and a national economic survey, providing a foundation for future research.

\end{itemize}

\section{Methods}
\subsection*{Overall Framework}
The \textit{\textbf{SocioVerse}} framework follows a structured pipeline to achieve realistic social simulation results, as shown in Figure~\ref{fig: framework}: (1) {\ul Social Environment} collects updated information and contextual knowledge. Within the simulation environment, (2) {\ul User Engine} aligns the simulated agents with target users, (3) {\ul Scenario Engine} aligns the interaction structure with diverse scenarios, and (4) {\ul Behavior Engine} aligns the behavioral pattern with real-world target groups.

\subsection{Social Environment}
\paragraph{Function} The social environment provides event-related context to align the simulation environment with real-world conditions. By integrating up-to-date events, social statistics, and preference content into LLM-based agents, it enhances the realism of the simulation and improve agent decision-making.


\paragraph{Components} The social environment should encompass as much real-world social, cultural, and technological context as possible. It can be broadly categorized into three types: social structural information, social dynamic information, and personalized context.

{\ul \textit{Social Structure:}} Social structural information provides agents with a rich knowledge base encompassing demographic distributions, cultural norms, urban infrastructures, and collective behavior patterns~\cite{wu2024pasum}. This data allows agents to behave in a way that aligns with the typical characteristics of their assigned demographic or geographic profile. For example, by incorporating regional dialect preferences, work-life habits, and common social values, the simulation can more accurately reflect public discourse trends, mobility behaviors, and economic interactions.

{\ul \textit{Social Dynamics:}} Social dynamics encompass time-sensitive content continuously generated in the real world, such as news events and policy changes. Typically, this engine maintains an up-to-date event base to continuously collect real-world event news from mainstream news, and all the news articles contain time stamps and event-related tags so that LLM-based agents can comb through the timeline of the events and react accordingly~\cite{mou2024unveiling}.

{\ul \textit{Personalized Context:}} In addition to the macro social environment, individuals also receive different personalized information feeds. Previous studies have explored that the recommendation system can enhance the behavior diversity of the agent~\cite{lyu2023llm, wang2023user, yang2024oasis}. Consequently, the preference content component constructs relevant posts and pushes them to agents according to their social interaction network and interesting topics.





\subsection{User Engine} 
\paragraph{Function}
The user engine aligns simulated agents with a rich set of real-world user samples, enabling the creation of complex target users within the simulation. 

\paragraph{Components} To support diverse user composition and effective user retrieval and description, the user engine incorporates a large user pool and a wide range of user labels.

{\ul \textit{User Pools:}} The user pool is designed to collect extensive digital footprints of individuals across social media platforms, enabling a more comprehensive characterization of real-world behavioral patterns and expression tendencies. To this end, we constructed a user pool covering a variety of social media platforms, including X\footnote{\url{https://x.com/}} and Rednote\footnote{\url{https://www.xiaohongshu.com/}}. Anomalous data, such as advertising and bot-generated content, is filtered by calculating the post frequency and average text similarity. The detailed procedure can be found in Appendix~\ref{app:data-clean}. We index users and construct a \textbf{user pool of 10 million users} based on the collected social media posts. Formally, we define user pool as: \(UserPool = \{U_i,P_i~|~ i\in \mathbb{S}\}\), where the \textit{i-th} user \(U_i\) derives from the collection of social media platforms \(\mathbb{S}\) with his/her related posts \(P_i = \{P_{i,1}, P_{i,2}, ... \}\). The statistical summary of the user pool is provided in Table~\ref{tab: user pool}.


\begin{table}[h]
\setlength{\abovecaptionskip}{0.2cm}
\setlength{\belowcaptionskip}{-0.2cm}
\centering
\resizebox{0.4\columnwidth}{!}{
\begin{tabular}{@{}lcc@{}}
\toprule[1.1pt]
Source  & \# Users & \# Posts \\ \midrule[1.1pt]
X & 1,006,517 & 30,195,510 \\
Rednote & 9,158,404 & 40,963,735 \\ \bottomrule[1.1pt]
\end{tabular}
}
\caption{Statistical summary of the 10M user pool.}
\label{tab: user pool}
\end{table}

{\ul \textit{User Labels:}} User labels refer to the tagging and description of users, which can be represented using discrete attributes or continuous representation. Demographic descriptions of users are the most commonly used form of labeling. However, they are often not directly accessible. Therefore, we designed a demographic annotation system to infer and label user attributes. 
The process begins with multiple LLMs serving as initial annotators, classifying users across various demographic dimensions.
Human annotators then evaluate and refine the LLM-generated labels, ensuring the reliability of the user tags dataset. The curated dataset is subsequently used to train demographic classifiers, enabling large-scale annotation in a cost-effective manner.
Specifically, we annotate users across 15 demographic dimensions:  \textit{age}, \textit{gender}, \textit{vocation}, \textit{race}, \textit{income}, \textit{education}, \textit{settlement type}, \textit{region}, \textit{employment}, \textit{marital status}, \textit{religious}, \textit{party}, \textit{ideology}, \textit{BigFive personality}, and \textit{hobbies}. Each attribute is inferred by a specialized classifier trained on the corresponding subset of the user tags dataset. See Appendix~\ref{app:demo} for further details.

\subsection{Scenario Engine}
\paragraph{Function}
The scenario engine aligns various simulation structures with real-world contexts based on specific task formulations and scenario types, and then scales individual simulations by sampling according to demographic distributions provided by the user engine.

\paragraph{Components} The scenario engine formulates a wide range of real-world social situations, which can be summarized as archetypal scenario templates, including questionnaires, in-depth interviews, behavior experiments, and social media interaction.

{\ul \textit{Questionnaire:}} The questionnaire scenario constructs the simulation in a 1-to-N manner, with one designed scale or questionnaire answered by multiple target users in a single round. This scenario is suitable for massive social investigation on specific topics, like election polls.

{\ul \textit{Indepth Interview:}} The in-depth interview scenario follows a 1-to-1 structure, where a simulated interviewer engages with an individual target user through multiple interaction rounds~\cite{park2024generative}. This iterative process allows for probing deeper into responses, clarifying ambiguities, and exploring underlying motivations. Such simulations are particularly useful for qualitative research on user experiences, psychological assessments, and exploratory studies where nuanced responses and detailed reasoning are essential.

{\ul \textit{Behavior Experiment:}} The behavior experiment scenario is typically conducted in a 1-to-N or N-to-N format, depending on whether individual or group interactions are being studied~\cite{Chandra:81, park2023generative}. Simulated users are exposed to controlled conditions where their behavioral responses are observed across multiple rounds of interaction. These simulations help researchers examine decision-making processes, social influences, and cognitive biases in various experimental setups, such as consumer behavior studies or cooperative game simulations.

{\ul \textit{Social Media Interaction:}} The social media interaction scenario adopts an N-to-N structure, where multiple simulated users engage in dynamic, multi-round exchanges in an online setting~\cite{liu2024skepticism}. This scenario captures real-time interactions, including content sharing, comment threads, and viral spread dynamics, allowing researchers to analyze public discourse, opinion shifts, and information diffusion on social platforms. It is particularly valuable for studying trends in misinformation, political discussions, and network-based influence propagation.

\subsection{Behavior Engine} 
\paragraph{Function}
The behavior engine aims to align the behaviors of the agents with that of real users. The behavior engine integrates user history and experience from the user engine, the interaction mechanism from the scenario engine and social context from the social environment to predict the behavior of each individual.

\paragraph{Components} To achieve credible behavior simulation, the behavior engine needs to provide a robust simulation foundation, including traditional agent-based models and a series of LLMs.

{\ul \textit{Traditional Agent-Based Modeling:}} Traditional agent-based modeling (ABM) relies on rule-based and mathematical models~\cite{schelling1969models,macal2009agent,jusup2022social,chuang2023computational,tang2024idea}, where interactions among agents are typically realized through the broadcasting of predefined values. These values are derived from heuristic functions or theoretical mathematical formulations. Traditional ABM approaches are highly scalable and computationally efficient, making them well-suited for simulating large populations, especially marginal users with relatively limited influence.

{\ul \textit{LLM-powered Agents:}} LLMs leverage their role-playing capabilities to simulate user-generated content, and the abilities can be activated through various methods~\cite{mou2024agentsensebenchmarkingsocialintelligence, yue2024synergistic, liu2024ai, zhang2024electionsim, sun2024identity, ye2025multi, zhang-etal-2024-somelvlm}. Specifically, the behavior engine can be powered by general LLMs, expert LLMs, and domain-specific LLMs. Through non-parametric prompting, powerful general LLMs (e.g., GPT series and Qwen series) can act in accordance with predefined user profiles. Expert and domain-specific LLMs are acquired through parametric training, e.g., continual pretraining, supervised fine-tuning, and reinforcement learning. When target users exhibit complex profiles and the simulation requires deep domain expertise, these models are leveraged to enhance the professionalism and accuracy of agent behaviors.

\section{Implementation of Specific Scenarios}
We implement three representative social simulation scenarios through the SocioVerse framework based on the implemented components: (a) {\ul presidential election prediction of America}, (b) {\ul breaking news feedback analysis}, and (c) {\ul national economic survey of China}. These scenarios respectively address political communication, journalistic dissemination, and socioeconomic domains, demonstrating the framework's generalizability through standardized implementation pipelines.

\begin{figure}[h]
    \centering
    \includegraphics[width=\linewidth]{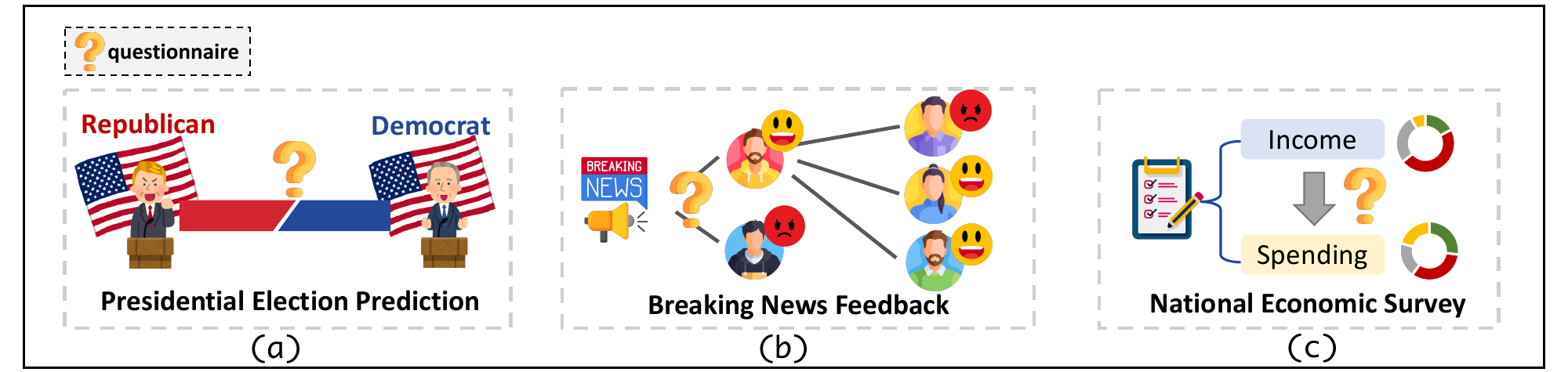}
    \caption{Illustration of three scenarios representing (a) presidential election prediction, (b) breaking news feedback, and (c) national economic survey.}
    \label{fig:scenario}
\end{figure}

\subsection{Presidential Election Prediction of America}
\paragraph{Task Description}
Presidential elections remain central to public engagement and party strategy formation~\cite{bartels1996uninformed,rosenstone1981forecasting}. This study analyzes methods for large-scale election simulation using LLMs through the U.S. presidential system's Electoral College framework. In this indirect voting system, citizens vote for state electors (allocated by congressional representation) who formally elect the president. Most states employ a winner-takes-all allocation of electoral votes to the statewide majority winner, with our modeling focused on predicting these \textbf{state-level} outcomes.

\paragraph{Target Group Distribution}
Extensive research has documented the influence of demographic factors on election outcomes~\cite{major2018threat, teixeira2009red}. We model U.S. demographic and ideological diversity through integrated Census Bureau (2022 voting/registration) and ANES (2020) data~\cite{anes2021}. This scenario incorporates 12 attributes from the user engine: socioeconomic (income, education, employment), geographic (region, area), and political (party, ideology) dimensions alongside demographic factors (age, gender, race, marital status, and religious status). Given available marginal distributions, we employ iterative proportional fitting (IPF) to synthesize agent populations, see Appendix~\ref{app:ipf}.

\paragraph{Questionnaire Design}
We design the presidential election questionnaire based on abundant polls conducted by various media and research institutes~\cite{barnett2023polls, keeter20212020pewresearch}, incorporating both significant issues and voter preferences. These elements are then optimized into proper forms for LLM-based agents by the scenario engine. The entire questionnaire can be found in Appendix~\ref{app:quesion-election}.

\paragraph{Evaluation Metric}
Two metrics are used to comprehensively compare the simulated election results to the real-world results. (1) Accuracy rate (Acc) is measured by calculating the proportion of states for which the election simulation results align with the actual result, serving as a coarse-grained evaluation metric. (2) Root Mean Square Error (RMSE) is measured by calculating the simulated vote share and the actual vote share for each state, which serves as a fine-grained evaluation metric.

\subsection{Breaking News Feedback}\label{subsec:news}
\paragraph{Task Description}
Journalism plays a crucial role in shaping public perception and opinion through agenda-setting, framing, and information dissemination~\cite{van2024revisiting, Gómez-Calderón_Ceballos_2024}. Online social media platforms have gradually replaced the influence of traditional paper media. When breaking news is released on social media platforms, its potential audience may hold different stances. We take \textit{the release of ChatGPT} as our target news to evaluate the accuracy and foreseeability of public attitudes.

\paragraph{Target Group Distribution}
We define all Rednote users in our pool as the universal set, identifying technology-interested users as the \textbf{potential audience set \(\mathbb{P}\)}, and those discussing ChatGPT via keyword matching as the \textbf{ground truth set \(\mathbb{G}\)}, with \(\mathbb{G}\subset\mathbb{P}\subset UserPool\). Context is limited to pre-news timeframes to prevent leakage. Using the potential audience distribution as prior, we sample agents with identical distribution sampling (IDS) as \(D_s=IDS(UserPool, \mathbb{P})\), see Appendix~\ref{app:id-sample}), considering demographics (\textit{gender, age, education, and consumption level}) during sampling the user pools. Based on this, the task is to compare the consistency between the agents' attitudes toward news and those of the users in the ground truth set.

\paragraph{Questionnaire Design}
We design the cognitive questionnaire using the ABC attitude model ({\ul A}ffect, {\ul B}ehavior, {\ul C}ognition)~\cite{liu2021public}, which outlines attitude formation as a hierarchy: cognition affects emotions, guiding behavior. Combined with a 5-point Likert scale~\cite{joshi2015likert}, the questionnaire covers six dimensions: public cognition (PC), perceived risks (PR), perceived benefits (PB), trust (TR), fairness (FA), and public acceptance (PA). See Appendix~\ref{app:question-news} for details.

\paragraph{Evaluation Metric}
Agents from both sets answer the questionnaire for paired responses. Two evaluation dimensions assess feedback: (1) Normalized RMSE (NRMSE) measures point-wise differences between simulated and ground truth answers across PC, PR, PB, TR, FA, and PA as value evaluations; (2) KL-divergence (KL-Div) compares the 6-dimensional answer distributions between groups as consistency evaluations.

\subsection{National Economic Survey of China}
\paragraph{Task Description}
Economic simulation is another crucial part of massive social simulations as it models resource distribution, market dynamics, and financial behaviors, providing insights into economic stability and policy impacts~\cite{dignum2020analysing, trimborn2020sabcemm}. By integrating economic factors with social interactions, it enhances the prediction of systemic outcomes, guiding decision-making in areas such as governance, urban planning, and crisis management. We follow a national economic survey conducted by the National Bureau of Statistics of China, which interviews Chinese citizens on their monthly spending given the average salary of each province in China.

\paragraph{Target Group Distribution}
The prior distribution is based on the methodology from the National Bureau of Statistics of China, which takes 160,000 families nationwide and calculates their incomes and spending as the national average statistics~\cite{NBS2023statistical}. We sample nationwide agents from our user pool proportionally according to their \textit{region} population and generate their \textit{income} distribution according to the regional average income~\cite{chinapopulation2023}. The detailed method can be referred to in Appendix~\ref{app:eco-dist}.

\paragraph{Questionnaire Design}\label{subsec:eco-quesitonnaire}
Spending details in China Statistical Yearbook 2024~\cite{Chinayearbook2024} are categorized into eight parts, i.e. \textit{food}, \textit{clothing}, \textit{housing}, \textit{daily necessities \& services}, \textit{communication \& transportation}, \textit{education \& entertainment}, \textit{healthcare}, and \textit{others}. Consequently, the questionnaire design covers the above categories with examples and uses segmented interval options in each question. The entire questionnaire can be referred to in Appendix~\ref{app:question-eco}.

\paragraph{Evaluation Metric}
Both value evaluation and distribution evaluation are involved in the national economic survey as well. (1) NRMSE of the nine categories is measured between the simulated results and official statistics. (2) KL-Div is measured by taking the 8-item spending as a distribution to evaluate the consistency between the simulation and the real world.

\section{Results}

\subsection{SocioVerse Can Support Diverse and Accurate Massive Social Simulations}

\begin{table}[h]
\setlength{\abovecaptionskip}{0.2cm}
\setlength{\belowcaptionskip}{-0.2cm}
\renewcommand{\arraystretch}{1.2}
\centering
\resizebox{\textwidth}{!}{
\begin{tabular}{@{}lcccccccc@{}}
\toprule[1.1pt]
Scenario & \# Agents & \# Demographics & Type & Sampling & Source & Language & \# Questions & Ground truth \\ \midrule[1.1pt]
PresElectPredict & 331,836 & 12 & label & IPF & X & EN & 49 & real world \\
BreakNewsFeed & 20,000 & 7 & label & IDS & rednote & ZH & 18 & calculated \\
NatEconSurvey & 16,000 & 9 & label+number & IDS & rednote & ZH & 17 & real world \\ \bottomrule[1.1pt]
\end{tabular}
}
\caption{Detail settings of three simulation scenarios, where PresElectPredict, BreakNewsFeed, and NatEconSurvey denote three simulations mentioned in the paper, respectively. IPF and IDS denote iterative proportional fitting and identical distribution sampling, see Appendix~\ref{app:sampling}.}
\label{tab:scenario}
\end{table}

\paragraph{Experiment Settings} We select powerful LLMs from different model families. For open-sourced models, we select Llama-3-70b-Instruct~\cite{dubey2024llama3}, Qwen2.5-72b-Instruct~\cite{yang2024qwen25}, DeepSeek-R1-671b~\cite{guo2025deepseek}, and DeepSeek-V3~\cite{liu2024deepseekv3}. For commercial models, we select GPT-4o\footnote{\texttt{gpt-4o-2024-08-06}}~\cite{OpenAI2024gpt4o} and GPT-4o-mini\footnote{\texttt{gpt-4o-mini-2024-07-18}}.

We compare the settings of all three scenarios for better understanding, which is shown in Table~\ref{tab:scenario}. As the Presidential Election Prediction covers a 1-in-1,000 sample of the U.S. population, GPT-4o is excluded from comparison due to cost constraints.
In terms of local model serving, Qwen2.5-72b-Instruct and Llama3-70b-Instruct models are both deployed on 8 NVIDIA RTX4090 GPUs via vLLM~\cite{kwon2023efficient}. We set max tokens to 2,048
for all models to enable chain-of-thoughts during the generation and the temperature is set to 0.7 to encourage diversity. Implementation details for user pool construction and demographics annotation can be found in Appendix~\ref{app:data-clean} and~\ref{app:demo}.

\begin{table}[h]
\setlength{\abovecaptionskip}{0.2cm}
\setlength{\belowcaptionskip}{-0.2cm}
\renewcommand{\arraystretch}{1.2}
\centering
\resizebox{\textwidth}{!}{
\begin{tabular}{@{}lcccccccccc@{}}
\toprule[1.1pt]
\multirow{3}{*}{\textbf{Model}} & \multicolumn{4}{c}{\textbf{PresElectPredict}} & \multicolumn{2}{c}{\textbf{BreakNewsFeed}} & \multicolumn{4}{c}{\textbf{NatEconSurvey}} \\ \cmidrule(r){2-5} \cmidrule(r){6-7} \cmidrule(r){8-11}
 & \multicolumn{2}{c}{\textit{Overall}} & \multicolumn{2}{c}{\textit{Battleground}} &  &  & \multicolumn{2}{c}{\textit{Overall}} & \multicolumn{2}{c}{\textit{Developed-region}} \\
 & Acc$\uparrow$ & RMSE$\downarrow$ & Acc$\uparrow$ & RMSE$\downarrow$ & KL-Div$\downarrow$ & RMSE$\downarrow$ & KL-Div$\downarrow$ & RMSE$\downarrow$ & KL-Div$\downarrow$ & RMSE$\downarrow$ \\ \midrule[1.1pt]
Llama3-70b & 0.843 & 0.064 & 0.733 & 0.045 & 0.668 & 0.199 & \textbf{0.016} & \textbf{0.026} & \textbf{0.013} & \textbf{0.025} \\
Qwen2.5-72b & \textbf{0.922} & \textbf{0.037} & 0.800 & \textbf{0.031} & \textbf{0.113} & 0.059 & 0.066 & 0.048 & 0.043 & 0.039 \\
DeepSeek-R1-671b & \textbackslash{} & \textbackslash{} & 0.670 & 0.065 & 0.383 & 0.082 & 0.059 & 0.045 & 0.045 & 0.036\\
DeepSeek-V3 & \textbf{0.922} & 0.046 & \textbf{0.867} & 0.041 & 0.263 & 0.072 & 0.035 & 0.036 & 0.023 & 0.030\\
GPT-4o-mini & \textbackslash{} & \textbackslash{} & 0.800 & 0.039 & 0.195 & 0.114 & 0.046 & 0.045 & 0.030 & 0.036 \\
GPT-4o & \textbackslash{} & \textbackslash{} & \textbackslash{} & \textbackslash{} & 0.196 & \textbf{0.055} & 0.062 & 0.051 & 0.036 & 0.038 \\ \bottomrule[1.1pt]
\end{tabular}
}
\caption{Overall results of the three scenarios, where subset \textit{Battleground} indicates battleground states in the U.S. in the presidential election and subset \textit{Developed-Region} indicates top-10 developed regions in China in terms of GDP.}
\label{tab:overall}
\end{table}

\paragraph{Results} The overall simulation results of the three scenarios are shown in Table~\ref{tab:overall}. We also report subset results for presidential election prediction and national economic survey.
\begin{itemize}
    \item \textbf{Presidential Election Prediction} We report the overall results and the battleground states' results separately. The prediction of battleground states is challenging even in the real world and thus becomes the focus during the election process. According to the results, GPT-4o-mini and Qwen2.5-72b show competitive performance both in Acc and RMSE. Typically, according to the winner-takes-all rule, \textbf{over 90\% state voting results are predicted correctly}, which means the simulation achieves a high-precision macroscopic reduction of the real-world election results. After the case study, we find that DeepSeek-R1-671b sometimes falls into overthinking, resulting in less accurate results.
    \item \textbf{Breaking News Feedback} The results measure the overall consistency of each model compared with the real-world users' reactions and attitudes. To this end, the performances of GPT-4o and Qwen2.5-72b are more aligned with real-world perspectives than other models in terms of KL-Div and NRMSE, respectively, and the following detailed analysis will demonstrate that \textbf{the models consistently capture and accurately predict public trends and opinions}.
    \item \textbf{National Economic Survey} We report the overall results and results for the top 10 regions by GDP (i.e., developed regions) separately. Generally, all the models closely align with real-world statistics. Llama3-70b shows a significant superiority over other models in the economic survey scenario and all the models perform better in the 1st-Region subset than overall. The results demonstrate that \textbf{individuals' spending habits can be accurately reproduced under the \textit{SocioVerse} framework, especially in developed regions}.
\end{itemize}

The overall results from both value evaluation and distribution evaluation of three simulations sufficiently prove that \textbf{\textit{SocioVerse}} can support diverse and accurate massive social simulations with a standard pipeline and minimal changes with human experts in the loop. However, the choice of underlying LLMs can affect simulation precision across different scenarios, highlighting the need for further study.

\subsection{Prior Distribution and Real-World Knowledge Can Enhance Simulation Accuracy in Presidential Election Predictions}

\begin{table}[h]
\setlength{\abovecaptionskip}{0.2cm}
\setlength{\belowcaptionskip}{-0.2cm}
\renewcommand{\arraystretch}{1.2}
\centering
\resizebox{0.5\columnwidth}{!}{
\begin{tabular}{@{}lcc@{}}
\toprule[1.1pt]
\textbf{Model} & \textbf{Acc$\uparrow$} & \textbf{RMSE$\downarrow$} \\ \midrule[1.1pt]
Llama3-70b & \textbf{0.733} & \textbf{0.045} \\
- w/o Knowledge & 0.533 & 0.051 \\
- w/o Knowledge \& Piror Distribution & 0.600 & 0.386 \\ \midrule
Qwen2.5-72b & \textbf{0.800} & \textbf{0.031} \\
- w/o Knowledge & \textbf{0.800} & 0.033 \\
- w/o Knowledge \& Piror Distribution & 0.600 & 0.370 \\ \midrule
GPT-4o-mini & \textbf{0.800} & \textbf{0.039} \\
- w/o Knowledge & \textbf{0.800} & 0.052 \\
- w/o Knowledge \& Piror Distribution & 0.667 & 0.323 \\ \bottomrule[1.1pt]
\end{tabular}
}
\caption{Ablation experiment results on the presidential election prediction simulation, where -w/o Knowledge denotes \textit{without real-world user knowledge} and -w/o Piror Distribution denotes \textit{using random demographics distribution}.}
\label{tab:ablation}
\end{table}

We conduct an ablation study on the presidential election prediction simulation to assess the impact of prior demographics distribution and real-world user knowledge. As shown in Table~\ref{tab:ablation}, prior demographics distribution significantly improves the accuracy of the simulation in both Acc and RMSE compared to random demographics distribution. Additionally, past posts from users on social media platforms improve the fine-grained performance, especially for Llama3-70b in Acc and all the models in RMSE. We can tell from the ablation study that \textbf{both prior distribution and real-world knowledge in the \textit{SocioVerse} pipeline are significant during the simulation}.

\subsection{Group Preference and Perspectives Can Be Well Reflected in Breaking News Feedback}

\begin{figure}[h]
    \centering
    \includegraphics[width=\linewidth]{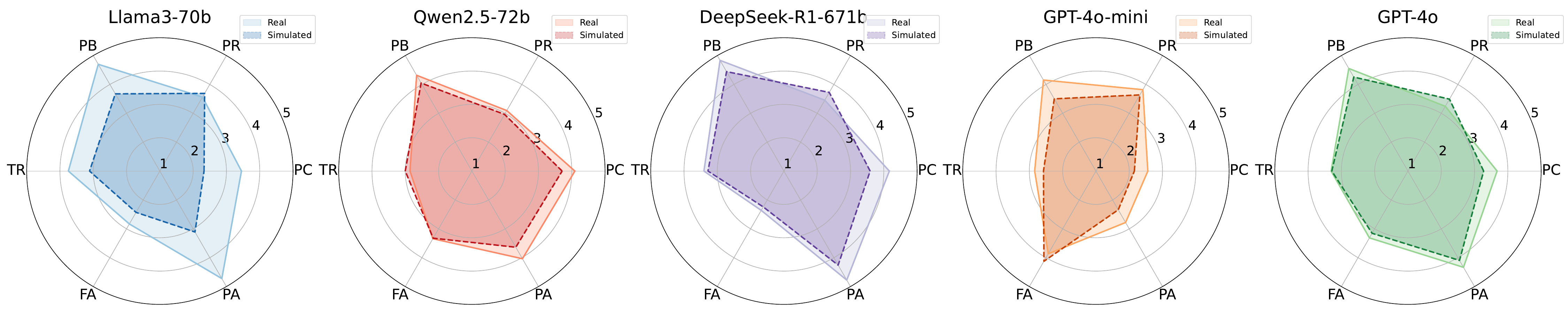}
    \caption{An illustration of the performances of the breaking news feedback simulation, where PC, PR, PB, TR, FA, and PA denote six dimensions from the Likert scale (see~\S\ref{subsec:news} questionnaire design), with 1-point standing for totally disagree and 5-point for totally agree.}
    \label{fig:news}
\end{figure}

During the Breaking News Feedback simulation, the core concern is whether the preferences and perspectives of the target group are well captured and reflected in the results. We reformulate the original questionnaire into the Likert 6-dimension scale ranging from 1 to 5 points, representing from totally disagree to totally agree. As the ground truth of the simulation is calculated by prompting LLM agents from the ground truth set, the \textit{simulated} and \textit{real} results are paired for each model, as shown in Figure~\ref{fig:news}. \textbf{All the models powered by the potential audience set during the simulation tend to behave consistently with the ground truth users.} However, Llama3-70b perform poorly with a larger gap between the \textit{simuated} and \textit{real} results than other models. GPT-4o-mini shows different attitudes in the fairness (FA) and public acceptance (PA) dimensions, which may be because the news is related to OpenAI.
Another trend indicates that, generally, \textbf{all the models perform more disagreeably in the \textit{simulated} results than the \textit{real} results}, which also underlines the potential risk of biases during the public opinion simulation. 

\subsection{The Capabilities of LLMs Vary in Different Domains in National Economic Survey}

\begin{table}[h]
\setlength{\abovecaptionskip}{0.2cm}
\setlength{\belowcaptionskip}{-0.2cm}
\renewcommand{\arraystretch}{1.2}
\centering
\resizebox{0.8\columnwidth}{!}{
\begin{tabular}{@{}lccccc@{}}
\toprule[1.1pt]
\textbf{Item} & \textbf{Llama3-70b} & \textbf{Qwen2.5-72b} & \textbf{GPT-4o-mini} & \textbf{GPT-4o} & \textbf{DeepSeek-R1} \\ \midrule[1.1pt]
Daily & \textbf{0.007} & {\ul 0.009} & \textbf{0.006} & {\ul 0.010} & \textbf{0.009} \\
Clothing & 0.012 & 0.015 & 0.019 & 0.015 & 0.015 \\
Transportation\_Communication & 0.016 & 0.020 & 0.027 & 0.023 & 0.017 \\
Education\_Entertainment & 0.018 & 0.022 & 0.024 & 0.017 & 0.022 \\
Medical & 0.023 & 0.062 & 0.041 & 0.057 & 0.060 \\
Food & 0.037 & 0.031 & 0.031 & 0.040 & 0.032 \\
Household & 0.052 & 0.110 & 0.107 & 0.120 & 0.102 \\ \midrule
Others & {\ul 0.008} & \textbf{0.008} & {\ul 0.010} & \textbf{0.005} & \textbf{0.009} \\ \bottomrule[1.1pt]
\end{tabular}
}
\caption{Detailed results on the national economic survey simulation reported in NRMSE, where the Item column indicates the components of spending. The best results are \textbf{bolded}; the second-best results are {\ul underlined}.}
\label{tab:eco-detail}
\end{table}

The simulation of the national economic survey covers 8 spending dimensions. The overall results in Table~\ref{tab:overall} show the average performance of these dimensions, while model performances among these dimensions can also vary. We calculate the averaged NRMSE of 31 regions on each spending level, as shown in Table~\ref{tab:eco-detail}. It is worth mentioning that all the models show high consistency. Eliminating the \textit{others} item, \textbf{all the models perform best on \textit{daily necessities} spending planning and worst on \textit{housing} spending}, which can reveal the LLM's preference on the economic decision-making and highlight the challenge in \textit{housing} spending strategy.

\section{Discussion}

In this study, we introduce a generalized social simulation framework \textbf{\textit{SocioVerse}} and evaluated its performance across three distinct real-world scenarios. Our findings indicate that state-of-the-art LLMs demonstrate a notable ability to simulate human responses in complex social contexts, although some gaps still remain between the simulated response and observed real-world outcomes. Therefore, future research may need to incorporate a broader range of scenarios and develop more fine-grained evaluations built upon the current analytic engine, to further explore and expand the boundaries of LLMs’ simulation capabilities. Such efforts could pave the way for establishing LLMs as comprehensive and reliable tools for large-scale social simulation.


We observed several key patterns across the simulations of the scenarios. First, incorporating demographic distributions and users' historical experiences significantly improved simulation accuracy. These findings highlight the importance of building a large, demographically rich user pool, complemented by a multi-dimensional user tagging system for more precise modeling of group-specific behaviors. Second, under consistent measurement protocols, LLMs produced broadly similar simulations of human attitudes and ideologies. However, certain models, such as GPT-4o-mini, showed notable inconsistencies, indicating that model-specific preferences or biases remain influential and warrant closer scrutiny in future work. Finally, we found that while LLMs perform well in simple daily scenarios, they underperform in complex situations requiring contextual knowledge, underscoring the need to align model behavior with real-world experiences and social contexts.


Notably, the current version has only implemented part of our framework, indicating significant potential for enhancing the accuracy and quality of social simulations. Future work can focus on refining each module for better collaboration, enabling the framework to achieve its full potential. For instance, the incorporation of the social environment can inject up-to-date knowledge into LLMs, enhancing the understanding of social dynamics. The scenario engine can not only provide survey-based simulation but also expand to diverse formats such as social interviews and free interactions. Additionally, further optimization of the general LLMs and expert LLMs adaptation in the behavior engine will enable better accommodation of complex target user groups, such as minority groups and individuals with special disabilities. The analysis engine can introduce an autonomous planning module to improve the overall credibility of simulation results.

Beyond the social simulation framework, our work underscores the potential to bridge the gap between autonomous AI systems and traditional social science, offering social scientists a seamless, cost-effective tool for conducting social experiments with minimal setup. Such tools not only help analyze and validate psychological and sociological theories or hypotheses, such as behavioral economics and social identity theory, but also assist in predicting large-scale social impacts like policy changes, social movements, or public health crises. By providing an efficient and scalable simulation environment, our framework is not just a research tool, but an experimental platform for exploring the dynamic changes and long-term trends of virtual societies, with the aim of becoming a realistic mapping for real-world societies.

\section*{Acknowledgement}
We would like to express our sincere gratitude to Professor Rongwei Chu and his research team for their invaluable support in this work. The project’s computational resources are supported by the CFFF platform of Fudan University.

\newpage
{
\small
\bibliographystyle{abbrv}
\bibliography{custom}
}

\newpage
\appendix
\section{Data Cleaning Details}\label{app:data-clean}

\subsection{Content Data Extraction}
We extract only post-related content on all the social media platforms to avoid violating privacy policies. Specifically, the data list on each platform is shown in Table~\ref{tab:app-data}.

\begin{table}[h]
\centering
\setlength{\abovecaptionskip}{0.2cm}
\setlength{\belowcaptionskip}{-0.2cm}
\resizebox{0.5\columnwidth}{!}{
\begin{tabular}{@{}lc@{}}
\toprule
Platform & Data list \\ \midrule
X & \begin{tabular}[c]{@{}c@{}}user ID, tweet, \#likes, \\ \#coments, \#retweets\end{tabular} \\
Rednote & user ID, notes, \#likes, \#comments \\ \bottomrule
\end{tabular}
}
\caption{Data list for each social media platform during the data collection.}
\label{tab:app-data}
\end{table}

\subsection{Abnormal Data Filtering}
We filter the abnormal data to guarantee the quality through text similarity calculation. Typically, all the textual content from the same user is calculated by means of the word repetition ratio. The threshold is set to 0.3. If the ratio surpasses the threshold, the user is considered likely to be a robot or advertising and will be filtered.

\newpage
\section{Demographics Annotation System}\label{app:demo}

\subsection{LLM Annotation}
To save costs, we first sample a subset of the user pool and employ multiple power LLMs for annotation. Due to the long time span of this work, users from different data sources in the user pool have used the powerful LLMs available at the time. For users derived from the X, GPT-4o\footnote{\texttt{gpt-4o-2024-08-06}}, Claude3.5-Sonnet\footnote{\texttt{claude-3-5-sonnet-20240620}}, and Gemini-1.5\footnote{gemini-1.5-pro} are employed. For users derived from the Rednote, GPT-4o, Cluade3.5-Sonnet, and Qwen2.5-72b are employed.

\subsection{Human Evaluation}
We employ 7 professional human annotators to verify the results annotated by LLMs. Typically, each annotator is required to re-annotate the demographic factors without the LLM labels. All the data are verified by at least 2 human annotators. The overall consistency between humans and LLMs is shown in Table~\ref{tab:app-demo-consis}.

\begin{table}[h]
\centering
\setlength{\abovecaptionskip}{0.2cm}
\setlength{\belowcaptionskip}{-0.2cm}
\resizebox{0.6\columnwidth}{!}{
\begin{tabular}{@{}lcc@{}}
\toprule
Models & Human (X) & Human (Rednote) \\ \midrule
GPT-4o & 0.905 & 0.723 \\
Claude3.5 & 0.901 & 0.659 \\
Gemini-1.5 & 0.713 & \textbackslash{} \\
Qwen2.5 & \textbackslash{} & 0.846 \\
Majority votes & \textbf{0.956} & \textbf{0.849} \\ \bottomrule
\end{tabular}
}
\caption{Human annotators' verification results. We report the consistency between humans and different LLMs.}
\label{tab:app-demo-consis}
\end{table}

\subsection{Classifier Training}
We take the majority-voted labels from different LLMs to construct the training dataset. Considering the difference in mainstream language used on different platforms, we employ LongFormer~\cite{beltagy2020longformer} for X data and employ Bert-base-chinese~\cite{devlin2019bert} for Rednote. The implementation details are shown in Table~\ref{tab:app-demo-train}.

\begin{table}[h]
\centering
\setlength{\abovecaptionskip}{0.2cm}
\setlength{\belowcaptionskip}{-0.2cm}
\resizebox{0.6\columnwidth}{!}{
\begin{tabular}{@{}lcc@{}}
\toprule
Params & LongFormer & Bert-base-chinese \\ \midrule
train\_size & 10,000 & 10,000 \\
\# classifiers & 5 & 4 \\
max\_tokens & 4096 & 512 \\
learning\_rate & 5e-5 & 5e-5 \\
batch\_size & 16 & 32 \\
optimizer & AdamW & AdamW \\
epochs & 3 & 10 \\
device & 8*4090 & 2*4090 \\ \bottomrule
\end{tabular}
}
\caption{Implementation details for demographic classifiers.}
\label{tab:app-demo-train}
\end{table}

We report the performances of demographic classifiers on each demographic factor in Table~\ref{tab:app-demo-res}.

\begin{table}[h]
\centering
\setlength{\abovecaptionskip}{0.2cm}
\setlength{\belowcaptionskip}{-0.2cm}
\resizebox{0.6\columnwidth}{!}{
\begin{tabular}{@{}lcccc@{}}
\toprule
\multirow{2}{*}{Demos} & \multicolumn{2}{c}{LongFormer} & \multicolumn{2}{c}{Bert-base-chinese} \\
 & Acc & F1 & Acc & F1 \\ \midrule
Gender & 0.875 & 0.904 & 0.926 & 0.958 \\
Age & 0.902 & 0.873 & 0.925 & 0.920 \\
Party & 0.849 & 0.846 & \textbackslash{} & \textbackslash{} \\
Ideology & 0.810 & 0.807 & \textbackslash{} & \textbackslash{} \\
Race & 0.779 & 0.768 & \textbackslash{} & \textbackslash{} \\
Consumption & \textbackslash{} & \textbackslash{} & 0.749 & 0.748 \\
Education & \textbackslash{} & \textbackslash{} & 0.954 & 0.975 \\ \bottomrule
\end{tabular}
}
\caption{Performance of demographic classifiers on test set.}
\label{tab:app-demo-res}
\end{table}

\subsection{Overall Distribution of the User Pool}
\begin{figure}[h]
    \centering
    \includegraphics[width=\linewidth]{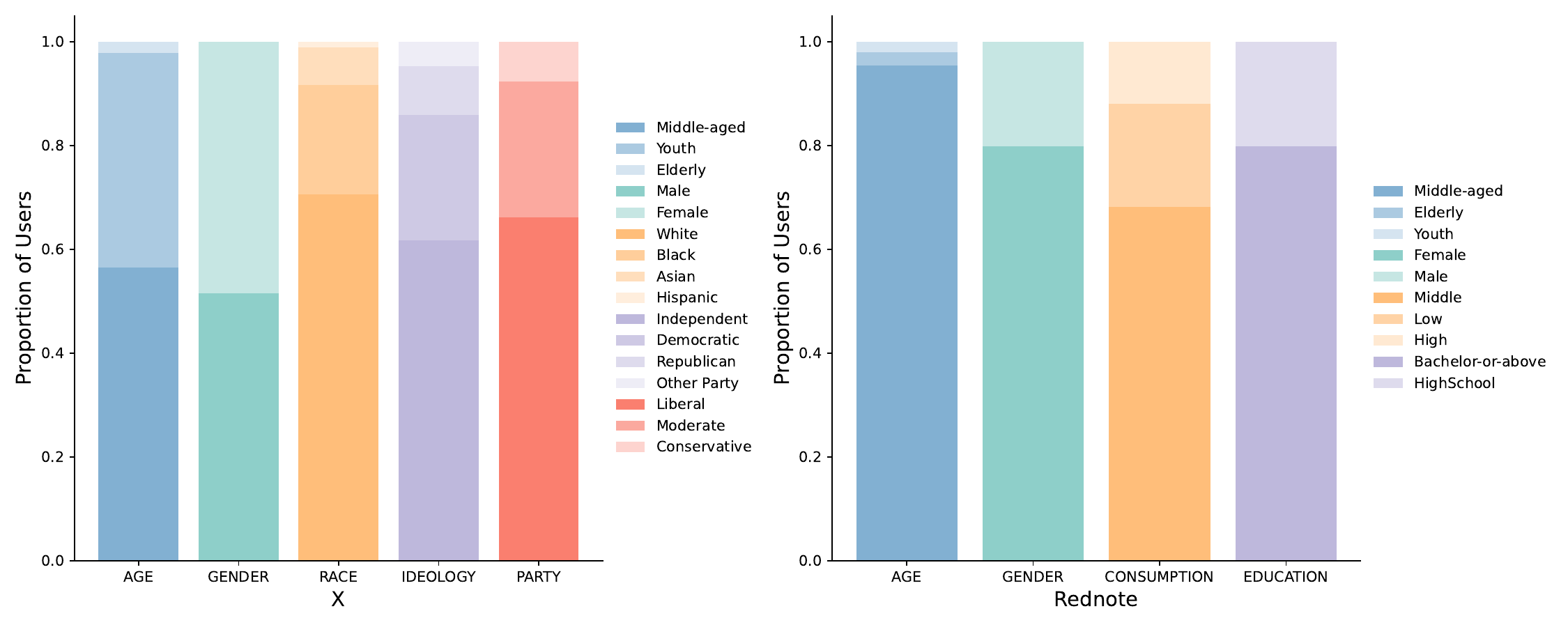}
    \caption{Demographic distribution on X and Rednote user pool.}
    \label{fig:app-demo-dist}
\end{figure}

We employ the demographic classifiers to annotate all of the users in the user pool, and the overall distributions are shown in Figure~\ref{fig:app-demo-dist}. For other demographics in specific simulations that are not considered in prior distribution, only users from the sampled user pool are annotated by the majority votes of LLMs.

\newpage
\section{Demographic Distribution Sampling Details}\label{app:sampling}

\subsection{Iterative Proportional Fitting}\label{app:ipf}
In our study, we follow the classical IPF method to construct the joint distribution of all the attributes in our simulation. Specifically, we start with a two-way table with individual components denoted as $x_{ij}$ and targeted estimation $\hat{x}_{ij}$. The targeted estimation $\hat{x}_{ij}$ satisfies $\Sigma_j \hat{x}_{ij} = v_i$ and $\Sigma_i \hat{x}_{ij} = w_j$. The iterations are specified as follows:

\begin{center}
\begin{minipage}{\columnwidth} 
Let $\hat{x}_{ij}^{(0)} = x_{ij}$. For $\alpha > 1$: 

\begin{equation}
 \hat{x}_{ij}^{(2\alpha-1)} = \dfrac{\hat{x}_{ij}^{(2\alpha - 2)}v_i}{\Sigma^J_{k=1}\hat{x}_{ij}^{(2\alpha - 2)}}   
\end{equation}

\begin{equation}
\hat{x}_{ij}^{(2\alpha)} = \dfrac{\hat{x}_{ij}^{(2\alpha - 1)}w_j}{\Sigma^I_{k=1}\hat{x}_{ij}^{(2\alpha - 1)}}  
\end{equation}

\end{minipage}
\end{center}

The iterations end when the estimated marginals are sufficiently close to the real marginals or when they stabilize without further convergence.

For the presidential election simulation, we implement the IPF algorithm for each state using five attributes: \textit{gender}, \textit{race}, \textit{age group}, \textit{ideology}, and \textit{partisanship}. In most cases, the algorithm does not converge, but the gaps between the estimated and actual marginals are less than 5\%, with 888 out of 918 marginals falling within this range. For the outliers, since IPF adjusts proportionally to the marginals, the overall ratio of marginals remains consistent. We then use the estimated joint distribution and marginals for our massive simulation.

\subsection{Identical Distribution Sampling}\label{app:id-sample}
Identical distribution sampling, also known as direct sampling, is applied when the joint distribution of multiple demographics is available. Given feature \(X\) and \(Y\), the joint distribution can be formulated as \(p(X, Y)\). Then, identical distribution sampling can be formulated as follows:
\begin{equation}
    (X_i,Y_i)\sim p(X,Y)\quad i=1,2,...,n
\end{equation}

For breaking news feedback simulations, as the ground truth set is directly from the Rednote, we can obtain all the users' demographics and calculate the joint distribution. Simultaneously, the scale of the user pool satisfies the direct sampling requirements.

\subsection{Prior Distribution of National Economic Survey}\label{app:eco-dist}
For the national economic survey distribution, only average income is available from the official data. As a result, we generate the prior income distribution at the regional level. The income distribution across different regions exhibits significant heterogeneity, often characterized by a right-skewed pattern. To model this distribution, we adopt a mixture distribution approach, combining a log-normal distribution for the majority of the population with a Pareto distribution for the high-income segment. This hybrid model captures both the bulk of wage earners and the long-tail effect observed in high-income groups.

Formally, let \(X\) denote an individual's wage. We assume that for the lower and middle-income groups \((X<x_{min})\), incomes follow a log-normal distribution:
\begin{center}
    \begin{minipage}{\columnwidth}
        \begin{equation}
            X \sim \log \operatorname{Normal}\left(\mu, \sigma^2\right)
        \end{equation}
    where
    \begin{small}
    \begin{equation}
        \mu=\ln \left(\frac{\mu_{\text {actual }}^2}{\sqrt{\sigma_{\text {actual }}^2+\mu_{\text {actual }}^2}}\right), \quad \sigma=\sqrt{\ln \left(1+\frac{\sigma_{\text {actual }}^2}{\mu_{\text {actual }}^2}\right)}
    \end{equation}
    \end{small}
    \end{minipage}
\end{center}

For the high-income group \((X\geq x_{min})\), wages follow a Pareto distribution:
\begin{equation}
    P(X \geq x)=C x^{-\alpha}, \quad x \geq x_{\min }
\end{equation}
where \(\alpha\) is the Pareto shape parameter determining the income concentration at the top. The proportion of individuals assigned to each distribution is governed by an empirical threshold ratio, typically set such that 90\% of the population follows the log-normal distribution while 10\% follows the Pareto distribution. This mixture approach provides a flexible yet robust framework for simulating realistic income distributions across diverse economic conditions. We set all the parameters empirically according to previous research and generate the income distribution for 31 regions in China (Hong Kong, Macao, and Taiwan are excluded).

\newpage
\section{Questionnaire Design Details}
We provide the questionnaires here for all three simulations.

\subsection{Questionnaire for Presidential Election Prediction}\label{app:quesion-election}
\renewcommand{\arraystretch}{1.25}

\begin{longtable}[htbp]{lp{12cm}}
    \toprule
    \textbf{Q01} & \textbf{Voting Behavior} \\
    \rowcolor{gray!10} Question &  ORDER OF MAJOR PARTY CANDIDATE NAMES \\
    Value Labels & 1. Democrat first / Republican second \newline
    2. Republican first / Democrat second \\
    \midrule
    
    \textbf{Q02} & \textbf{Social Security} \\
    \rowcolor{gray!10} Question &  Next I am going to read you a list of federal programs. For each one, I would like you to tell me whether you would like to see spending increased, decreased, or kept the same. \newline
    What about Social Security? Should federal spending on Social Security be increased, decreased, or kept the same? \\
    Value Labels & -2. DK/RF \newline
1. Increased \newline
2. Decreased \newline
3. Kept the same  \\
    \midrule

    \textbf{Q03} & \textbf{Education} \\
    \rowcolor{gray!10} Question & What about public schools? Should federal spending on public schools be increased, decreased, or kept the same?\\
    Value Labels &  -2. DK/RF \newline
1. Increased \newline
2. Decreased \newline
3. Kept the same
 \\
    \midrule

    \textbf{Q04} & \textbf{Immigration} \\
    \rowcolor{gray!10} Question & What about tightening border security to prevent illegal immigration? Should federal spending on tightening border security to prevent illegal immigration be increased, decreased, or kept the same? \\
    Value Labels &  -2. DK/RF \newline
1. Increased \newline
2. Decreased \newline
3. Kept the same
 \\
    \midrule

    \textbf{Q05} & \textbf{Criminal Justice} \\
    \rowcolor{gray!10} Question & What about dealing with crime? Should federal spending on dealing with crime be increased, decreased, or kept the same? \\
    Value Labels &  -2. DK/RF \newline
1. Increased \newline
2. Decreased \newline
3. Kept the same
\\
    \midrule

    \textbf{Q06} & \textbf{Social Welfare} \\
    \rowcolor{gray!10} Question & What about welfare programs? Should federal spending on welfare programs be increased, decreased, or kept the same?\\
    Value Labels &  -2. DK/RF \newline
1. Increased \newline
2. Decreased \newline
3. Kept the same
 \\
    \midrule

    \textbf{Q07} & \textbf{Infrastructure} \\
    \rowcolor{gray!10} Question & What about building and repairing highways? Should federal spending on building and repairing highways be increased, decreased, or kept the same? \\
    Value Labels &  -2. DK/RF \newline
1. Increased \newline
2. Decreased \newline
3. Kept the same \\
    \midrule

    \textbf{Q08} & \textbf{Aid to Poor} \\
    \rowcolor{gray!10} Question & What about aid to the poor? Should federal spending on aid to the poor be increased, decreased, or kept the same? \\
    Value Labels &  -2. DK/RF \newline
1. Increased \newline
2. Decreased \newline
3. Kept the same \\
    \midrule

    \textbf{Q09} & \textbf{Environment} \\
    \rowcolor{gray!10} Question & What about protecting the environment? Should federal spending on protecting the environment be increased, decreased, or kept the same? \\
    Value Labels & -2. DK/RF \newline
1. Increased \newline
2. Decreased \newline
3. Kept the same  \\
    \midrule

    \textbf{Q10} & \textbf{Government} \\
    \rowcolor{gray!10} Question & How much do you feel that having elections makes the government pay attention to what the people think? \\
    Value Labels & -2. DK/RF \newline
1. A good deal \newline
2. Some \newline
3. Not much
  \\
    \midrule

    \textbf{Q11} & \textbf{Economy} \\
    \rowcolor{gray!10} Question & Which party do you think would do a better job of handling the nation’s economy?\\
    Value Labels &  -2. DK/RF \newline
1. Democrats would do a better job \newline
2. Not much difference between them \newline
3. Republicans would do a better job \\
    \midrule

    \textbf{Q12} & \textbf{Health Care} \\
    \rowcolor{gray!10} Question & Which party do you think would do a better job of handling health care?\\
    Value Labels & -2. DK/RF \newline
1. Democrats would do a better job \newline
2. Not much difference between them \newline
3. Republicans would do a better job 
   \\
    \midrule

    \textbf{Q13} & \textbf{Immigration} \\
    \rowcolor{gray!10} Question & Which party do you think would do a better job of handling immigration? \\
    Value Labels & -2. DK/RF \newline
1. Democrats would do a better job \newline
2. Not much difference between them \newline
3. Republicans would do a better job 
  \\
    \midrule

    \textbf{Q14} & \textbf{Taxes} \\
    \rowcolor{gray!10} Question & Which party do you think would do a better job of handling taxes?\\
    Value Labels & -2. DK/RF \newline
1. Democrats would do a better job \newline
2. Not much difference between them \newline
3. Republicans would do a better job   \\
    \midrule

    \textbf{Q15} & \textbf{Environment} \\
    \rowcolor{gray!10} Question & Which party do you think would do a better job of handling the environment? \\
    Value Labels & -2. DK/RF \newline
1. Democrats would do a better job \newline
2. Not much difference between them \newline
3. Republicans would do a better job 
  \\
    \midrule

    \textbf{Q16} & \textbf{Education} \\
    \rowcolor{gray!10} Question & Some people think the government should provide fewer services even in areas such as health and education in order to reduce spending. \newline
Other people feel it is important for the government to provide many more services even if it means an increase in spending. \newline
And, of course, some people have a neutral position. \newline
Which of the following best describes your view?
\\
    Value Labels & -2. DK/RF \newline
1. Government should provide fewer services \newline
2. Neutral\newline
3. Government should provide more services 
  \\
    \midrule

    \textbf{Q17} & \textbf{Defense} \\
    \rowcolor{gray!10} Question & Some people believe that we should spend less money for defense. \newline
Others feel that defense spending should be increased. \newline
And, of course, some people have a neutral position. \newline
Which of the following best describes your view?
\\
    Value Labels & -2. DK/RF \newline
1. Decrease defense spending \newline
2. Neutral\newline
3. Increase defense spending
  \\
    \midrule

    \textbf{Q18} & \textbf{Health Care} \\
    \rowcolor{gray!10} Question & There is much concern about the rapid rise in medical and hospital costs. \newline
Some people feel there should be a government insurance plan which would cover all medical and hospital expenses for everyone. \newline
Others feel that all medical expenses should be paid by individuals through private insurance plans like Blue Cross or other company paid plans. \newline
And, of course, some people have a neutral position. \newline
Which of the following best describes your view?
\\
    Value Labels & -2. DK/RF \newline
1. Government insurance plan \newline
2. Neutral \newline
3. Private insurance plan 
  \\
    \midrule

    \textbf{Q19} & \textbf{Social Welfare} \\
    \rowcolor{gray!10} Question & Some people feel the government in Washington should see to it that every person has a job and a good standard of living. \newline
Others think the government should just let each person get ahead on their own. \newline
And, of course, some people have a neutral position.\newline
Which of the following best describes your view?
\\
    Value Labels & -2. DK/RF \newline
1. Government should see to jobs and standard of living \newline
2. Neutral\newline
3. Government should let each person get ahead on own 
  \\
    \midrule

    \textbf{Q20} & \textbf{Aid to Blacks} \\
    \rowcolor{gray!10} Question & Some people feel that the government in Washington should make every effort to improve the social and economic position of blacks. \newline
Others feel that the government should not make any special effort to help blacks because they should help themselves. \newline
And, of course, some people have a neutral position.\newline
Which of the following best describes your view?
\\
    Value Labels & -2. DK/RF \newline
1. Government should help blacks \newline
2. Neutral\newline
3. Blacks should help themselves 
  \\
    \midrule

    \textbf{Q21} & \textbf{Environment} \\
    \rowcolor{gray!10} Question & Some people think we need much tougher government regulations on business in order to protect the environment. \newline
Others think that current regulations to protect the environment are already too much of a burden on business. \newline
And, of course, some people have a neutral position.\newline
Which of the following best describes your view?
\\
    Value Labels & -2. DK/RF \newline
1. Tougher regulations on business needed to protect environment \newline
2. Neutral\newline
3. Regulations to protect environment already too much a burden on business 
  \\
    \midrule

    \textbf{Q22} & \textbf{Abortion} \\
    \rowcolor{gray!10} Question & Would you be pleased, upset, or neither pleased nor upset if the Supreme Court reduced abortion rights? \\
    Value Labels & -2. DK/RF \newline
1. Pleased \newline
2. Upset \newline
3. Neither pleased nor upset
  \\
    \midrule

    \textbf{Q23} & \textbf{Criminal Justice} \\
    \rowcolor{gray!10} Question & Do you favor or oppose the death penalty for persons convicted of murder? \\
    Value Labels & -2. DK/RF \newline
1. Favor \newline
2. Oppose 
  \\
    \midrule

    \textbf{Q24} & \textbf{US Position in World} \\
    \rowcolor{gray!10} Question & Do you agree or disagree with this statement: `This country would be better off if we just stayed home and did not concern ourselves with problems in other parts of the world.' \\
    Value Labels & -2. DK/RF \newline
1. Agree \newline
2. Disagree  \\
    \midrule

    \textbf{Q25} & \textbf{US Position in World} \\
    \rowcolor{gray!10} Question & How willing should the United States be to use military force to solve international problems? \\
    Value Labels & -2. DK/RF \newline
1. Willing \newline
2. Moderately willing \newline 
3. Not willing
  \\
    \midrule

    \textbf{Q26} & \textbf{Inequality} \\
    \rowcolor{gray!10} Question & Do you think the difference in incomes between rich people and poor people in the United States today is larger, smaller, or about the same as it was 20 years ago? \\
    Value Labels & -2. DK/RF \newline 
1. Larger \newline 
2. Smaller \newline 
3. About the same
  \\
    \midrule

    \textbf{Q27} & \textbf{Environment} \\
    \rowcolor{gray!10} Question & Do you think the federal government should be doing more about rising temperatures, should be doing less, or is it currently doing the right amount? \\
    Value Labels &  -2. DK/RF \newline
1. Should be doing more \newline
2. Should be doing less \newline
3. Is currently doing the right amount
 \\
    \midrule

    \textbf{Q28} & \textbf{Parental Leave} \\
    \rowcolor{gray!10} Question & Do you favor, oppose, or neither favor nor oppose requiring employers to offer paid leave to parents of new children? \\
    Value Labels & -2. DK/RF \newline
1. Favor \newline
2. Oppose \newline
3. Neither favor nor oppose   \\
    \midrule

    \textbf{Q29} & \textbf{LGBTQ+ Rights} \\
    \rowcolor{gray!10} Question & Do you think business owners who provide wedding-related services should be allowed to refuse services to same-sex couples if same-sex marriage violates their religious beliefs, or do you think business owners should be required to provide services regardless of a couple’s sexual orientation? \\
    Value Labels & -2. DK/RF \newline
1. Should be allowed to refuse \newline
2. Should be required to provide services
  \\
    \midrule

    \textbf{Q30} & \textbf{LGBTQ+ Rights} \\
    \rowcolor{gray!10} Question & Should transgender people - that is, people who identify themselves as the sex or gender different from the one they were born as - have to use the bathrooms of the gender they were born as, or should they be allowed to use the bathrooms of their identified gender?\\
    Value Labels & -2. DK/RF \newline
1. Have to use the bathrooms of the gender they were born as \newline
2. Be allowed to use the bathrooms of their identified gender
  \\
    \midrule

    \textbf{Q31} & \textbf{LGBTQ+ Rights} \\
    \rowcolor{gray!10} Question & Do you favor or oppose laws to protect gays and lesbians against job discrimination? \\
    Value Labels &  -2. DK/RF \newline
1. Favor \newline
2. Oppose 
 \\
    \midrule

    \textbf{Q32} & \textbf{LGBTQ+ Rights} \\
    \rowcolor{gray!10} Question & Do you think gay or lesbian couples should be legally permitted to adopt children? \\
    Value Labels & -2. DK/RF \newline
1. Yes \newline
2. No  \\
    \midrule

    \textbf{Q33} & \textbf{LGBTQ+ Rights} \\
    \rowcolor{gray!10} Question & Which comes closest to your view? You can just tell me the number of your choice. \\
    Value Labels & -2. DK/RF
1. Gay and lesbian couples should be allowed to legally marry \newline
2. Gay and lesbian couples should be allowed to form civil unions but not legally marry  \newline
3. There should be no legal recognition of gay or lesbian couples’ relationship
  \\
    \midrule

    \textbf{Q34} & \textbf{Immigration} \\
    \rowcolor{gray!10} Question & Some people have proposed that the U.S. Constitution should be changed so that the children of unauthorized immigrants do not automatically get citizenship if they are born in this country. \newline
Do you favor, oppose, or neither favor nor oppose this proposal?\\
    Value Labels & -2. DK/RF \newline
1. Favor \newline
2. Oppose \newline
3. Neither favor nor oppose
  \\
    \midrule

    \textbf{Q35} & \textbf{Immigration} \\
    \rowcolor{gray!10} Question & What should happen to immigrants who were brought to the U.S. illegally as children and have lived here for at least 10 years and graduated high school here? Should they be sent back where they came from, or should they be allowed to live and work in the United States? \\
    Value Labels & -2. DK/RF\newline
1. Should be sent back where they came from \newline
2. Should be allowed to live and work in the US
  \\
    \midrule

    \textbf{Q36} & \textbf{Immigration} \\
    \rowcolor{gray!10} Question & Do you favor, oppose, or neither favor nor oppose building a wall on the U.S. border with Mexico?\\
    Value Labels & -2. DK/RF \newline
1. Favor \newline
2. Oppose \newline
3. Neither favor nor oppose  \\
    \midrule

    \textbf{Q37} & \textbf{Unrest} \\
    \rowcolor{gray!10} Question & During the past few months, would you say that most of the actions taken by protestors to get the things they want have been violent, or have most of these actions by protesters been peaceful, or have these actions been equally violent and peaceful? \\
    Value Labels & -2. DK/RF \newline
1. Mostly violent \newline
2. Mostly peaceful \newline
3. Equally violent and peaceful
  \\
    \midrule

    \textbf{Q38} & \textbf{Government} \\
    \rowcolor{gray!10} Question & Do you think it is better when one party controls both the presidency and Congress, better when control is split between the Democrats and Republicans, or doesn’t it matter?\\
    Value Labels &  -2. DK/RF \newline
1. Better when one party controls both \newline
2. Better when control is split \newline
3. It doesn’t matter
 \\
    \midrule

    \textbf{Q39} & \textbf{Government} \\
    \rowcolor{gray!10} Question & Would you say the government is pretty much run by a few big interests looking out for themselves or that it is run for the benefit of all the people? \\
    Value Labels & -2. DK/RF\newline
1. Run by a few big interests \newline
2. For the benefit of all the people 
  \\
    \midrule

    \textbf{Q40} & \textbf{Government} \\
    \rowcolor{gray!10} Question & Do you think that people in government waste a lot of the money we pay in taxes, waste some of it, or don’t waste very much of it?\\
    Value Labels & -2. DK/RF \newline
1. Waste a lot \newline
2. Waste some \newline
3. Don’t waste very much
  \\
    \midrule

    \textbf{Q41} & \textbf{Election Integrity} \\
    \rowcolor{gray!10} Question & Do you favor, oppose, or neither favor nor oppose allowing convicted felons to vote once they complete their sentence? \\
    Value Labels &  -2. DK/RF \newline
1. Favor \newline
2. Oppose \newline
3. Neither favor nor oppose
 \\
    \midrule

    \textbf{Q42} & \textbf{Democratic Norms} \\
    \rowcolor{gray!10} Question & How important is it that news organizations are free to criticize political leaders? \\
    Value Labels &  -2. DK/RF \newline
1. Not important\newline
2. Moderately important\newline
3. Important
 \\
    \midrule

    \textbf{Q43} & \textbf{Democratic Norms} \\
    \rowcolor{gray!10} Question & How important is it that the executive, legislative, and judicial branches of government keep one another from having too much power? \\
    Value Labels & -2. DK/RF \newline
1. Not important\newline
2. Moderately important\newline
3. Important  \\
    \midrule

    \textbf{Q44} & \textbf{Democratic Norms} \\
    \rowcolor{gray!10} Question & How important is it that elected officials face serious consequences if they engage in misconduct?\\
    Value Labels & -2. DK/RF \newline
1. Not important\newline
2. Moderately important\newline
3. Important  \\
    \midrule

    \textbf{Q45} & \textbf{Democratic Norms} \\
    \rowcolor{gray!10} Question & How important is it that people agree on basic facts even if they disagree politically? \\
    Value Labels & -2. DK/RF \newline
1. Not important\newline
2. Moderately important\newline
3. Important  \\
    \midrule

    \textbf{Q46} & \textbf{Democratic Norms} \\
    \rowcolor{gray!10} Question & Would it be helpful, harmful, or neither helpful nor harmful if U.S. presidents could work on the country’s problems without paying attention to what Congress and the courts say?\\
    Value Labels &  -2. DK/RF \newline
1. Helpful \newline
2. Harmful \newline
3. Neither helpful nor harmful
 \\
    \midrule

    \textbf{Q47} & \textbf{Democratic Norms} \\
    \rowcolor{gray!10} Question & Do you favor, oppose, or neither favor nor oppose elected officials restricting journalists’ access to information about government decision-making? \\
    Value Labels & -2. DK/RF \newline
1. Favor \newline
2. Oppose \newline
3. Neither favor nor oppose
  \\
    \midrule

    \textbf{Q48} & \textbf{Gender Resentment} \\
    \rowcolor{gray!10} Question & `Many women interpret innocent remarks or acts as being sexist.'\newline
Do you agree, neither agree nor disagree, or disagree with this statement?\\
    Value Labels & -2. DK/RF/technical error \newline
1. Agree \newline
2. Neither agree nor disagree \newline
3. Disagree
  \\
    \midrule

    \textbf{Q49} & \textbf{Gender Resentment} \\
    \rowcolor{gray!10} Question & `Women seek to gain power by getting control over men.' \newline
Do you agree, neither agree nor disagree, or disagree with this statement?\\
    Value Labels &  -2. DK/RF/technical error \newline
1. Agree \newline
2. Neither agree nor disagree \newline
3. Disagree \\
    \bottomrule

\end{longtable}

\subsection{Questionnaire for Breaking News Feedback}\label{app:question-news}
\renewcommand{\arraystretch}{1.25}

\begin{longtable}[htbp]{lp{12cm}}

    \toprule
    
    \textbf{Q01} & \textbf{Public Cognition (PC)} \\
    \rowcolor{gray!10} Question &  I have heard of ChatGPT. \\
    Value Labels & 1. Disagree \newline
    2. Partially disagree\newline
    3. Neutral\newline
    4. Partially agree\newline
    5. Agree\\
    \midrule

    \textbf{Q02} & \textbf{Public Cognition (PC)} \\
    \rowcolor{gray!10} Question &  Many people around me use ChatGPT. \\
    Value Labels & 1. Disagree \newline
    2. Partially disagree\newline
    3. Neutral\newline
    4. Partially agree\newline
    5. Agree\\
    \midrule

    \textbf{Q03} & \textbf{Public Cognition (PC)} \\
    \rowcolor{gray!10} Question &  I have a deep understanding of ChatGPT's functions and applications. \\
    Value Labels & 1. Disagree \newline
    2. Partially disagree\newline
    3. Neutral\newline
    4. Partially agree\newline
    5. Agree\\
    \midrule

    \textbf{Q04} & \textbf{Perceived Risks (PR)} \\
    \rowcolor{gray!10} Question &  ChatGPT may lead to the widespread dissemination of false information. \\
    Value Labels & 1. Disagree \newline
    2. Partially disagree\newline
    3. Neutral\newline
    4. Partially agree\newline
    5. Agree\\
    \midrule

    \textbf{Q05} & \textbf{Perceived Risks (PR)} \\
    \rowcolor{gray!10} Question &  ChatGPT may reduce human thinking ability and creativity. \\
    Value Labels & 1. Disagree \newline
    2. Partially disagree\newline
    3. Neutral\newline
    4. Partially agree\newline
    5. Agree\\
    \midrule

    \textbf{Q06} & \textbf{Perceived Risks (PR)} \\
    \rowcolor{gray!10} Question &  The development of ChatGPT may replace certain jobs, and I am deeply concerned about this. \\
    Value Labels & 1. Disagree \newline
    2. Partially disagree\newline
    3. Neutral\newline
    4. Partially agree\newline
    5. Agree\\
    \midrule

    \textbf{Q07} & \textbf{Perceived Benefits (PB)} \\
    \rowcolor{gray!10} Question &  ChatGPT will definitely improve my work and study efficiency. \\
    Value Labels & 1. Disagree \newline
    2. Partially disagree\newline
    3. Neutral\newline
    4. Partially agree\newline
    5. Agree\\
    \midrule

    \textbf{Q08} & \textbf{Perceived Benefits (PB)} \\
    \rowcolor{gray!10} Question &  ChatGPT helps broaden my knowledge and provides me with new perspectives and ideas. \\
    Value Labels & 1. Disagree \newline
    2. Partially disagree\newline
    3. Neutral\newline
    4. Partially agree\newline
    5. Agree\\
    \midrule

    \textbf{Q09} & \textbf{Perceived Benefits (PB)} \\
    \rowcolor{gray!10} Question &  ChatGPT promotes technological innovation and development in related fields. \\
    Value Labels & 1. Disagree \newline
    2. Partially disagree\newline
    3. Neutral\newline
    4. Partially agree\newline
    5. Agree\\
    \midrule

    \textbf{Q10} & \textbf{Trust (TR)} \\
    \rowcolor{gray!10} Question &  I fully trust the team developing ChatGPT to manage and guide its development responsibly. \\
    Value Labels & 1. Disagree \newline
    2. Partially disagree\newline
    3. Neutral\newline
    4. Partially agree\newline
    5. Agree\\
    \midrule

    \textbf{Q11} & \textbf{Trust (TR)} \\
    \rowcolor{gray!10} Question &  I have strong confidence in the accuracy and reliability of the information generated by ChatGPT. \\
    Value Labels & 1. Disagree \newline
    2. Partially disagree\newline
    3. Neutral\newline
    4. Partially agree\newline
    5. Agree\\
    \midrule

    \textbf{Q12} & \textbf{Trust (TR)} \\
    \rowcolor{gray!10} Question &  I believe that the future application of ChatGPT will be effectively regulated. \\
    Value Labels & 1. Disagree \newline
    2. Partially disagree\newline
    3. Neutral\newline
    4. Partially agree\newline
    5. Agree\\
    \midrule

    \textbf{Q13} & \textbf{Fairness (FA)} \\
    \rowcolor{gray!10} Question &  The opportunities to use ChatGPT are distributed fairly among different groups of people. \\
    Value Labels & 1. Disagree \newline
    2. Partially disagree\newline
    3. Neutral\newline
    4. Partially agree\newline
    5. Agree\\
    \midrule

    \textbf{Q14} & \textbf{Fairness (FA)} \\
    \rowcolor{gray!10} Question &  I find the distribution of benefits brought by ChatGPT to be fair. \\
    Value Labels & 1. Disagree \newline
    2. Partially disagree\newline
    3. Neutral\newline
    4. Partially agree\newline
    5. Agree\\
    \midrule

    \textbf{Q15} & \textbf{Fairness (FA)} \\
    \rowcolor{gray!10} Question &  I believe that the decision-making process for the development and promotion of ChatGPT is fully transparent and adequately reflects public interests. \\
    Value Labels & 1. Disagree \newline
    2. Partially disagree\newline
    3. Neutral\newline
    4. Partially agree\newline
    5. Agree\\
    \midrule

    \textbf{Q16} & \textbf{Public Acceptance (PA)} \\
    \rowcolor{gray!10} Question &  Overall, I strongly welcome the emergence of ChatGPT. \\
    Value Labels & 1. Disagree \newline
    2. Partially disagree\newline
    3. Neutral\newline
    4. Partially agree\newline
    5. Agree\\
    \midrule

    \textbf{Q17} & \textbf{Public Acceptance (PA)} \\
    \rowcolor{gray!10} Question &  I am definitely willing to use ChatGPT in my work or studies. \\
    Value Labels & 1. Disagree \newline
    2. Partially disagree\newline
    3. Neutral\newline
    4. Partially agree\newline
    5. Agree\\
    \midrule

    \textbf{Q18} & \textbf{Public Acceptance (PA)} \\
    \rowcolor{gray!10} Question &  I strongly support increased investment in the research and development of AI technologies like ChatGPT. \\
    Value Labels & 1. Disagree \newline
    2. Partially disagree\newline
    3. Neutral\newline
    4. Partially agree\newline
    5. Agree\\
    \bottomrule

\end{longtable}

\subsection{Questionnaire for National Economic Survey}\label{app:question-eco}
\renewcommand{\arraystretch}{1.25}

\begin{longtable}[htbp]{lp{12cm}}

    \toprule
    
    \textbf{Q01} & \textbf{Food} \\
    \rowcolor{gray!10} Question &  What is your average monthly expenditure on food (including dining out)? (Unit: CNY) \\
    Value Labels & A. Below 500 CNY \newline
    B. 501-650 CNY\newline
    C. 651-800 CNY\newline
    D. 801-1000 CNY\newline
    E. Above 1000 CNY\\
    \midrule

    \textbf{Q02} & \textbf{Food} \\
    \rowcolor{gray!10} Question &  Do you think your current spending on food, tobacco, and alcohol is too high relative to your income? \\
    Value Labels & A. Yes\newline B. No\newline C. Acceptable\\
    \midrule

    \textbf{Q03} & \textbf{Clothing} \\
    \rowcolor{gray!10} Question &  What is your average monthly expenditure on clothing (including apparel, shoes, and accessories)? (Unit: CNY) \\
    Value Labels & A. Below 50 CNY\newline B. 51-100 CNY\newline C. 101-150 CNY\newline D. 151-200 CNY\newline E. Above 200 CNY\\
    \midrule

    \textbf{Q04} & \textbf{Clothing} \\
    \rowcolor{gray!10} Question &  How much economic pressure do you feel from clothing expenses? \\
    Value Labels & A. Very low, almost no pressure\newline B. Moderate, some pressure but manageable\newline C. High, requires careful spending\newline D. Very high, affects spending in other areas\\
    \midrule

    \textbf{Q05} & \textbf{Household} \\
    \rowcolor{gray!10} Question &  What is your average monthly housing expenditure? (Including rent, mortgage, property fees, maintenance, etc.) (Unit: CNY) \\
    Value Labels & A. Below 200 CNY\newline B. 201-500 CNY\newline C. 501-800 CNY\newline D. 801-1200 CNY\newline E. Above 1200 CNY\\
    \midrule

    \textbf{Q06} & \textbf{Household} \\
    \rowcolor{gray!10} Question &  What percentage of your monthly income is spent on housing? (Including rent, mortgage, property fees, maintenance, etc.) \\
    Value Labels & A. Below 10\%\newline B. 10\%-20\%\newline C. 21\%-30\%\newline D. 31\%-40\%\newline E. Above 40\%\\
    \midrule

    \textbf{Q07} & \textbf{Daily Service} \\
    \rowcolor{gray!10} Question &  What is your average monthly expenditure on daily necessities (personal care, household items, cleaning supplies, etc.) and services (housekeeping, repairs, beauty, pet services, etc.)? (Unit: CNY) \\
    Value Labels & A. Below 80 CNY\newline B. 81-120 CNY\newline C. 121-160 CNY\newline D. 161-200 CNY\newline E. Above 200 CNY\\
    \midrule

    \textbf{Q08} & \textbf{Transportation \& Communication} \\
    \rowcolor{gray!10} Question &  What is your average monthly expenditure on transportation (public transport, taxis, fuel, parking, etc.) and communication (mobile and internet fees)? (Unit: CNY) \\
    Value Labels & A. Below 200 CNY\newline B. 201-300 CNY\newline C. 301-400 CNY\newline D. 401-500 CNY\newline E. Above 500 CNY\\
    \midrule

    \textbf{Q09} & \textbf{Education \& Entertainment} \\
    \rowcolor{gray!10} Question &  What is your average monthly expenditure on education (tuition, training, books, etc.) and cultural entertainment (movies, performances, games, fitness, cultural activities, etc.)? (Unit: CNY) \\
    Value Labels & A. Below 100 CNY\newline B. 101-200 CNY\newline C. 201-300 CNY\newline D. 301-400 CNY\newline E. Above 400 CNY\\
    \midrule

    \textbf{Q10} & \textbf{Education \& Entertainment} \\
    \rowcolor{gray!10} Question &  Can you easily afford your current education, cultural, and entertainment expenses? \\
    Value Labels & A. Yes, spending does not affect other areas\newline B. Barely, needs some control\newline C. Not really, affects other expenditures\newline D. No, it creates significant financial pressure\\
    \midrule

    \textbf{Q11} & \textbf{Medical} \\
    \rowcolor{gray!10} Question &  What is your average monthly expenditure on healthcare (medications, medical services, health management, etc.)? (Unit: CNY) \\
    Value Labels & A. Below 100 CNY\newline B. 101-200 CNY\newline C. 201-300 CNY\newline D. 301-400 CNY\newline E. Above 400 CNY\\
    \midrule

    \textbf{Q12} & \textbf{Medical} \\
    \rowcolor{gray!10} Question &  Have you purchased private medical or health insurance for yourself or your family? \\
    Value Labels & A. Yes\newline B. Not yet, but planning to\newline C. No, and no plans to\\
    \midrule

    \textbf{Q13} & \textbf{Others} \\
    \rowcolor{gray!10} Question &  Besides food, clothing, housing, daily necessities and services, transportation, education, culture, and healthcare, what is your average monthly expenditure on other areas (e.g., hobbies, charitable donations, investment, etc.)? (Unit: CNY) \\
    Value Labels & A. Below 30 CNY\newline B. 31-60 CNY\newline C. 61-90 CNY\newline D. 91-120 CNY\newline E. Above 120 CNY\\
    \midrule

    \textbf{Q14} & \textbf{Overall} \\
    \rowcolor{gray!10} Question &  How would you evaluate the impact of your current consumption level on your household (or personal) financial situation? \\
    Value Labels & A. Comfortable, can moderately increase spending\newline B. Average, can maintain current spending\newline C. Tight, need to control or reduce spending\newline D. Very tight, affects quality of life\\
    \midrule

    \textbf{Q15} & \textbf{Overall} \\
    \rowcolor{gray!10} Question &  Do you feel that your consumption pressure is too high relative to your income level? \\
    Value Labels & A. Yes\newline B. No\newline C. Not sure\\
    \midrule

    \textbf{Q16} & \textbf{Overall} \\
    \rowcolor{gray!10} Question &  If your income increases, which consumption areas would you most like to expand or improve? (Multiple choices allowed) \\
    Value Labels & A. Food and alcohol\newline B. Clothing\newline C. Housing\newline D. Daily necessities and services\newline E. Transportation and communication\newline F. Education, culture, and entertainment\newline G. Healthcare\newline H. Other goods and services\\
    \midrule

    \textbf{Q17} & \textbf{Overall} \\
    \rowcolor{gray!10} Question &  What is your consumption expectation for the next six months to a year? \\
    Value Labels & A. Will continue to increase\newline B. Will remain roughly the same\newline C. Will moderately decrease\newline D. Uncertain\\
    \bottomrule

\end{longtable}

\end{document}